\documentclass[11pt]{article}

\usepackage[final]{acl}

\usepackage{times}
\usepackage{latexsym}
\usepackage{amsmath}
\usepackage{amssymb}
\usepackage{multicol}
\usepackage{booktabs}
\usepackage[dvipsnames]{xcolor}
\usepackage[T1]{fontenc}
\usepackage[utf8]{inputenc}

\usepackage{microtype}

\usepackage{inconsolata}

\usepackage{graphicx}

\title{Detecting Overflow in Compressed Token Representations for Retrieval-Augmented Generation}
\author{
 \textbf{Julia Belikova\textsuperscript{1,2}},
 \textbf{Danila Rozhevskii\textsuperscript{1}},
 \textbf{Dennis Svirin\textsuperscript{1,4}},
 \\
 \textbf{Konstantin Polev\textsuperscript{2}},
 \textbf{and Alexander Panchenko\textsuperscript{1,3}}
\\
\textsuperscript{1}Skoltech,
\textsuperscript{2}Sber AI Lab,
\textsuperscript{3}AIRI
\\
\textsuperscript{4}Institute for Information Transmission Problems of the Russian Academy of Sciences
                            \\
 \small{
   \textbf{Correspondence:} \href{mailto:julia.belikova@skol.tech}{\{julia.belikova, a.panchenko\}@skol.tech}
 }
}

\begin{document}
\maketitle

\begin{abstract}

Efficient long-context processing remains a crucial challenge for contemporary large language models (LLMs), especially in resource-constrained environments. Soft compression architectures promise to extend effective context length by replacing long token sequences with smaller sets of learned \emph{compressed tokens}. Yet, the limits of compressibility -- and when compression begins to erase task-relevant content -- remain underexplored. In this paper, we define \emph{token overflow} as a regime in which compressed representations no longer contain sufficient information to answer a given query, and propose a methodology to characterize and detect it.
In the xRAG soft-compression setting, we find that query-agnostic saturation statistics reliably separate compressed from uncompressed token representations, providing a practical tool for identifying compressed tokens but showing limited overflow detection capability. Lightweight probing classifiers over both query and context xRAG representations detect overflow with 0.72 AUC-ROC on average on HotpotQA, SQuADv2, and TriviaQA datasets, demonstrating that incorporating query information improves detection performance. These results advance from query-independent diagnostics to query-aware detectors, enabling low-cost pre-LLM gating to mitigate compression-induced errors.

\end{abstract}

\section{Introduction}

Large language models (LLMs) remain computationally constrained when processing long contexts, even as architectural advances and extended context windows become widely available~\citep{vaswani2023attentionneed,liu2023lostmiddlelanguagemodels}. In retrieval-augmented generation (RAG), this limitation is particularly acute: retrieved evidence must be aggressively compressed or truncated, creating a tension between efficiency and faithfulness~\citep{lewis2021retrievalaugmentedgenerationknowledgeintensivenlp,aushev-etal-2025-ragulator}. Soft compression architectures address this by mapping long contexts into dense vectors that can be directly consumed by the model, dramatically reducing token count while preserving global semantics~\citep{liao2025hardsofthybridcontext}.

% However, the same mechanism that enables extreme compression also introduces a critical failure mode. As more information is packed into fixed-dimensional vectors, these \emph{compressed tokens} can become \emph{saturated}: they cease to carry task-relevant signal and instead behave like noise, silently degrading downstream performance. Recent work on trainable tokens shows that individual embeddings have substantial theoretical capacity, but also that practical limits depend strongly on architecture, training, and input complexity~\citep{kuratov2025cramming}. Yet current compression systems are typically evaluated only via end-task metrics, offering little insight into \emph{when} compressed tokens cross from informative to overflowed states.

However, the same mechanism that enables extreme compression also introduces a critical failure mode. As more information is packed into a fixed-dimensional \emph{compressed token}, its representation can enter token overflow: it no longer carries sufficient task-relevant signal for the query and effectively behaves like noise, silently degrading downstream performance. Recent work on trainable tokens shows that individual embeddings have substantial theoretical capacity, but also that practical limits depend strongly on architecture, training, and input complexity~\citep{kuratov2025cramming}. Yet, current compression systems are typically evaluated only via end-task metrics, offering little insight into \emph{when} a single compressed token crosses from informative to overflowed states.

This paper investigates token overflow in soft compression architectures. We ask: \textbf{(RQ1)} How can we characterize overflow in compressed representations? \textbf{(RQ2)} Can overflow be detected efficiently, without full LLM inference, using lightweight diagnostics? \textbf{(RQ3)} Is overflow detectable from compressed tokens alone, or does it require modeling query-context interactions?

To address these questions, we:
\begin{itemize}
\item formalize \textbf{token overflow} and propose a methodology advancing from query-independent to query-aware detection approaches;
\item demonstrate that \textbf{saturation statistics} reliably distinguish compressed tokens from standard tokens, providing a practical tool for identifying compressed representations, but show limited overflow detection capability;
\item show that \textbf{attention patterns} during generation provide moderate overflow signal but require LLM forward passes;
\item develop \textbf{learned probing classifiers} operating on joint query-context representations that achieve strong overflow detection without LLM inference (Table~\ref{tab:main_results}), showing that incorporating query information improves detection performance;
% \item present empirical results on SQuADv2 establishing that overflow detection improves systematically as query specificity is incorporated, with supervised contrastive learning outperforming standard probing by 5-6\%.
\end{itemize}

Although our experiments focus on the xRAG architecture, the methodology is general and we expect it to yield similar results when applied to other setups. The source code is publicly available online\footnote{\url{https://github.com/s-nlp/overflow-detection}}.

\section{Related Work}
\paragraph{Long-context modeling and compression} Efficient long-context processing has been tackled through architectural changes~\citep{beltagy2020longformerlongdocumenttransformer,zaheer2021bigbirdtransformerslonger,dai2019transformerxlattentivelanguagemodels} and explicit compression. Context compression is systematized into \emph{hard}, \emph{soft}, and \emph{hybrid} paradigms~\citep{liao2025hardsofthybridcontext}: hard compression selects token subsets with strict information bottlenecks; soft compression maps contexts into dense vectors accessible via attention; hybrid methods combine both approaches. Our work analyzes the \emph{failure modes} of soft compression, asking when compressed vectors fail to carry useful task information.
\paragraph{Soft compression in RAG} Retrieval-augmented generation (RAG) frameworks extend LLMs with external corpora~\citep{lewis2021retrievalaugmentedgenerationknowledgeintensivenlp}, motivating compression of retrieved passages. Several soft compression methods have been proposed: AutoCompressors~\cite{chevalier-etal-2023-adapting} learn summary vectors by training the model to reconstruct compressed context through attention; ICAE~\cite{ge2024incontextautoencodercontextcompression} employs in-context autoencoding to compress sequences into memory slots with combined reconstruction and language modeling objectives ($\sim$4$\times$ compression, 1\% parameters).
We focus our experiments on xRAG~\citep{xRAG}, utilizing it not merely as a baseline, but as a representative \emph{projector-based} compression paradigm. Unlike autoencoder-based methods that compress context via complex recurrence or reconstruction objectives, xRAG treats dense retrieval embeddings as a distinct modality. It employs a lightweight projector to map these embeddings directly into the LLM's input space. This architectural choice isolates the compression mechanism from the complexities of extensive parameter fine-tuning ($<$0.1\% parameters), allowing us to study the interactions between the projector and the frozen LLM in a controlled setting. By decoupling the retrieval representation from the generative process, xRAG provides clean access to pre- and post-projection states, making it an ideal testbed for analyzing signal degradation and token saturation without the confounders of end-to-end model adaptation.
% \paragraph{Gisting and learned prompt compression}
% Gisting~\citep{DBLP:conf/nips/Mu0G23} is a soft compression approach that trains LLMs to condense prompts into a small set of learned \emph{gist tokens} via attention masking, forcing downstream computation to rely on compressed prefix activations. Unlike projector-based methods such as xRAG~\citep{xRAG}, gisting performs compression entirely within the LLM’s representation space and amortizes compression through instruction tuning. While gisting shows that few dense tokens can preserve task behavior, prior work largely evaluates downstream quality, leaving open when gist representations saturate and lose task-relevant information—an issue closely related to the overflow phenomenon we study.
\paragraph{Motivation for overflow detection}
Detecting \emph{information overflow} -- where input complexity exceeds compressed token capacity -- is critical for optimizing RAG pipelines. It enables \emph{adaptive chunking}, allowing systems to dynamically resize input segments based on semantic density rather than arbitrary fixed lengths. Furthermore, \emph{early overflow detection} facilitates computational pruning: identifying and discarding saturated representations immediately after projection prevents wasteful LLM inference on degraded context.
Despite progress across methods, most evaluations treat compressed vectors as black boxes and focus on downstream metrics~\citep{ge2024incontextautoencodercontextcompression,xRAG}. Recent work shows single vectors can theoretically encode thousands of tokens, yet \emph{practical} capacity depends on architecture and complexity~\citep{kuratov2025cramming}. In contrast, we \emph{operationalize} capacity limits through overflow detection in xRAG, advancing from query-independent saturation statistics to query-aware learned probing.

\section{Methodology}

Our goal is to characterize and detect \emph{token overflow} in soft compression architectures across tasks and context regimes. We focus on xRAG-style compressors attached to frozen LLM backbones, studying how compressed token properties change as context complexity increases and downstream quality degrades. We employ a spectrum of detection approaches with increasing query-awareness: from query-agnostic saturation statistics, through query-conditioned attention patterns, to fully query-aware learned probing classifiers.

Our methodology is motivated by the observation that the same compressed representation may be sufficient for one query but overflowed for another. This motivates our approach, advancing from query-independent to query-aware detection:

\begin{enumerate}
\item \textbf{Context complexity and saturation statistics} (query-agnostic): Measure intrinsic properties of compressed representations independent of any query -- useful for identifying and characterizing compressed tokens.
\item \textbf{Attention features} (query-conditioned): Analyze how the LLM utilizes compressed tokens during generation for a specific query -- captures behavioral signals but requires LLM forward passes.
\item \textbf{Learned probing} (query-aware): Train classifiers on joint query-context representations to detect overflow in embedding space -- achieves strong detection without LLM inference.
\end{enumerate}

This approach allows us to evaluate whether overflow detection improves as query information is incorporated, while identifying the most efficient deployment strategy.

\subsection{Problem Setup}

Let $\mathcal{M}$ be a frozen LLM and $\mathcal{C}$ a soft compression module (e.g., xRAG’s modality-fusion compressor) that maps an input sequence of $n$ tokens with embeddings $\mathbf{X} \in \mathbb{R}^{n \times d}$ to $k \ll n$ compressed tokens $\mathbf{C} = \mathcal{C}(\mathbf{X}) \in \mathbb{R}^{k \times d}$. The compressed tokens are then injected into $\mathcal{M}$ (e.g., as extra prefix tokens or interleaved context) and used to solve a downstream task such as extractive QA.

Given an input instance $i$ with original context $x_i$, question $q_i$, and gold output $y_i$, we define task performance under compression, $\mathcal{T}_i(\mathbf{C}_i)$, as a scalar metric (e.g., F1, EM, or ROUGE), indicating whether the generated answer is judged correct. We compare it to a reference performance $\mathcal{T}_i^{\text{ref}}$ obtained from either (i) an uncompressed baseline (full context within the model’s window), or (ii) a lightly compressed setting where degradation is empirically negligible.

We define an \emph{overflow} state for instance $i$ as:
\begin{equation}
\mathcal{O}_i = \mathbf{1}\!\left(\mathcal{T}_i^{\text{ref}} = 1 \;\land\; \mathcal{T}_i(\mathbf{C}_i) = 0\right).
\end{equation}
 Our objective is to (a) understand how compressed representations differ between overflow and non-overflow regimes, and (b) learn detectors that can predict overflow from representations alone, without recomputing $\mathcal{T}_i$.

More generally, this formulation can be extended by defining overflow via a degradation threshold,
\begin{equation}
\mathcal{T}_i^{\text{ref}} - \mathcal{T}_i(\mathbf{C}_i) \ge \epsilon,
\end{equation}
where $\epsilon$ is task-dependent. Exploring such threshold-based criteria with alternative evaluation functions is a promising direction for future work.

\subsection{Context Complexity Measures}\label{subsec:context}

For each input context $x_i$ (or aggregated retrieved context), we compute a set of \emph{context complexity} measures intended to approximate how “hard” the context is to compress:

\begin{itemize}
\item \textbf{Context length} $N_{ctx}$: the number of tokens in the original, uncompressed context before any truncation. This is the simplest proxy for potential compression pressure and directly correlates with computational cost.
\item \textbf{Language-model perplexity} $\text{PPL}_i$: the average per-token negative log-likelihood under the base LLM (without compression), which captures how predictable the context is given the model’s training distribution. Higher perplexity indicates linguistically or semantically atypical content.
\item \textbf{Statistical compressibility} $R_i$: the compression ratio achieved by a standard lossless compressor (e.g., gzip or LZMA) on the raw text. We define $R_i = \frac{|x_i|_{\text{bytes}}}{|\text{zip}(x_i)|_{\text{bytes}}}$, where larger values indicate more redundancy and thus higher statistical compressibility.
\end{itemize}

These metrics allow us to analyze how overflow correlates with raw length, lexical predictability, and sequence-level redundancy (compressibility).

\subsection{Token Saturation Statistics}

We quantify saturation at the level of compressed tokens and their propagated hidden states. For each compressed token vector $\mathbf{c} \in \mathbb{R}^d$ and its corresponding hidden states $\mathbf{h}^{(\ell)}$ at layer $\ell$, we compute the following statistics.

\paragraph{Hoyer's sparsity} Hoyer’s index~\citep{hoyer2004nonnegativematrixfactorizationsparseness} measures how concentrated a vector’s energy is in a few dimensions:
\begin{equation}
H(\mathbf{v}) = \frac{\sqrt{d} - \frac{\lVert \mathbf{v} \rVert_1}{\lVert \mathbf{v} \rVert_2}}{\sqrt{d} - 1}.
\end{equation}
It ranges from 0 (all components equal) to 1 (only one non-zero component). Informative compressed tokens are hypothesized to exhibit higher sparsity (structured, selective activations), while overflowed tokens tend towards low sparsity (flat, noise-like patterns).

\paragraph{Spectral entropy} We apply a discrete cosine transform (DCT) to $\mathbf{v}$ and treat the normalized squared magnitudes as an energy distribution $p$ over frequency components. The spectral entropy is defined as folowing:
\begin{equation}
S(\mathbf{v}) = - \sum_{i=1}^{d} p_i \log p_i,\quad
p_i = \frac{|\text{DCT}(\mathbf{v})_i|^2}{\lVert \text{DCT}(\mathbf{v}) \rVert_2^2}.
\end{equation}
Low entropy corresponds to concentrated energy (structured signals), whereas near-maximum entropy indicates white-noise–like spectra.

\paragraph{Kurtosis} We compute the excess kurtosis of the entries of $\mathbf{v}$:
\begin{equation}
K(\mathbf{v}) = \frac{\mathbb{E}[(v_j - \mu)^4]}{\sigma^4} - 3,
\end{equation}
where $\mu$ and $\sigma$ are the mean and standard deviation across dimensions. Heavy-tailed distributions (positive kurtosis) suggest a few large, informative coordinates, while overflowed tokens are expected to approach Gaussian-like behavior ($K \approx 0$).

\subsection{Attention Features: Query-conditioned Overflow Signals}\label{subsec:attention}

While saturation statistics measure intrinsic token properties, they ignore how the LLM actually \emph{uses} compressed tokens during generation. To capture this behavioral dimension, we extract attention-based features that quantify the model's reliance on xRAG tokens when processing a specific query.

For each instance, we perform a forward pass through the LLM with both the query and compressed context, extracting attention weights $\mathbf{A} \in \mathbb{R}^{L \times H \times T \times T}$ across all layers $L$, heads $H$, and sequence positions $T$. We compute:

\paragraph{Mean attention to xRAG tokens} For each layer $\ell$ and head $h$, we measure the average attention mass directed to compressed token positions:
\begin{equation}
\bar{a}_{\text{xRAG}}^{(\ell, h)} = \frac{1}{|T_q|} \sum_{i \in T_q} \sum_{j \in T_{\text{xRAG}}} A_{i,j}^{(\ell, h)},
\end{equation}
where $T_q$ denotes query token positions and $T_{\text{xRAG}}$ denotes xRAG token positions. We aggregate across layers and heads to obtain instance-level statistics: mean, max, min, and standard deviation of attention to xRAG tokens.

\paragraph{Attention ratios} To contextualize xRAG attention, we compute ratios comparing attention to compressed versus uncompressed tokens:
\begin{equation}
r_{\text{xRAG/non-xRAG}} = \frac{\bar{a}_{\text{xRAG}}}{\bar{a}_{\text{non-xRAG}}}.
\end{equation}
This ratio isolates whether the model preferentially attends to compressed representations or relies more heavily on other context.

\paragraph{Attention entropy} For each query position $i$, we compute the entropy of its attention distribution over all positions:
\begin{equation}
Ent_i = -\sum_{j=1}^{T} A_{i,j} \log A_{i,j}.
\end{equation}
High entropy indicates diffuse attention (potentially signaling uncertainty or lack of relevant information), while low entropy indicates focused attention to specific tokens~\cite{rykov-etal-2025-smurfcat}.

\subsection{Overflow Detection Methods}

We evaluate overflow detection through two complementary approaches that span a spectrum from interpretable to representational methods. First, we test \textbf{feature-based classification} using explicit, hand-crafted features to determine whether overflow manifests in interpretable, low-dimensional signals. Second, we develop \textbf{learned probing classifiers} that operate directly on high-dimensional query and context embedding vectors.

\paragraph{Feature-based classification} We aggregate the hand-crafted features described in \S\ref{subsec:context}--\S\ref{subsec:attention} (context complexity, saturation statistics, attention patterns) and train a \textbf{logistic regression classifier} implemented in scikit-learn\footnote{\url{https://scikit-learn.org}}. All hyperparameters are detailed in Appendix~\ref{app:hyperparameters}.

\paragraph{Learned probing on vector representations} While feature-based methods offer interpretability, they may fail to capture complex interactions between query and context that manifest in the geometry of representation spaces. We therefore develop \textbf{learned probing classifiers} that operate directly on joint query-context representations. Our hypothesis is that overflow detection requires modeling alignment patterns in shared representation space.

\paragraph{Representation extraction} 
For each instance $i$ with query $q_i$ and context $x_i$, we extract embeddings at multiple stages:

\begin{itemize}
    \item \textit{Query representations}: $q_i^{\text{preproj}} \in \mathbb{R}^{d_{\text{ret}}}$ (retriever embedding), $q_i^{\text{postproj}} \in \mathbb{R}^{d_{\text{LLM}}}$ (after projection), $q_i^{\text{mid}}, q_i^{\text{last}} \in \mathbb{R}^{d_{\text{LLM}}}$ (hidden states from intermediate and final LLM layers).
    \item \textit{Context representations}: $x_i^{\text{preproj}} \in \mathbb{R}^{d_{\text{ret}}}$ (retriever embedding), $x_i^{\text{postproj}} \in \mathbb{R}^{d_{\text{LLM}}}$ (compressed token after projection), $x_i^{\text{mid}}, x_i^{\text{last}} \in \mathbb{R}^{d_{\text{LLM}}}$ (hidden states from intermediate and final layers).
\end{itemize}

We construct \textbf{joint feature vectors} by concatenating query and context representations at matched or complementary stages:
\begin{equation}
\phi_i = [\, x_i^{(s_c)} ; q_i^{(s_q)} \,], 
\end{equation}
where $s_c, s_q \in \{\text{preproj}, \text{postproj}, \text{mid}, \text{last}\}$ denote the extraction stage. Our primary experiments use projection-stage representations (preproj, postproj) which are available immediately after encoding without requiring LLM forward passes. Following prior work demonstrating that intermediate transformer layers encode complementary information useful for interpretation tasks~\citep{wang2024androidshallucination, belikova2025metamodels}, we additionally evaluate multi-layer representations (including $\text{mid}, \text{last}$) to assess the efficiency-accuracy trade-off.

\paragraph{Classifier architectures}
To systematically assess the role of model capacity and training objectives in overflow detection, we evaluate three neural probe architectures:

\begin{itemize}
    \item \textit{Linear Probe}: A single linear transformation applied to the joint feature vector $\phi_i$. This minimal architecture tests whether overflow is linearly separable in the concatenated representation space.
    \item \textit{MLP Probe}: A two-layer feedforward network with one hidden layer, introducing nonlinear feature interactions while maintaining computational efficiency.
    \item \textit{MLP Probe with Supervised Contrastive Learning (SCL)}: An MLP trained with a hybrid objective that combines standard binary cross-entropy with a supervised contrastive term~\citep{khosla2021supervisedcontrastivelearning}. This architecture explicitly structures the representation space by encouraging same-class instances to cluster while pushing apart opposite-class instances.
\end{itemize}

For the SCL probe, we minimize the combined objective
\begin{equation}
\mathcal{L} = \mathcal{L}_{\mathrm{BCE}} + \lambda\,\mathcal{L}_{\mathrm{SCL}},
\end{equation}
where $\mathcal{L}_{\mathrm{BCE}}$ provides direct classification supervision, while $\mathcal{L}_{\mathrm{SCL}}$ imposes metric constraints by maximizing cosine similarity between same-label pairs and minimizing it between different-label pairs in the learned representation space.

\section{Results}

\subsection{Experimental Setup}

As a preliminary study, we use the xRAG-7B model\footnote{\url{https://hf.co/Hannibal046/xrag-7b}} as the base LLM and SFR-Embedding-Mistral\footnote{\url{https://hf.co/Salesforce/SFR-Embedding-Mistral}} as the retriever embedding model for all experiments. We focus on three extractive question answering datasets: \textbf{SQuADv2}~\citep{rajpurkar-etal-2018-know}, a context-based QA benchmark over Wikipedia passages; \textbf{TriviaQA}~\citep{joshi2017triviaqalargescaledistantly}, a large-scale reading comprehension dataset with independently collected evidence documents; and \textbf{HotpotQA}~\citep{DBLP:conf/emnlp/Yang0ZBCSM18}, a multi-hop reasoning dataset requiring information synthesis across multiple paragraphs. All three datasets were part of the xRAG compression module's training data, providing a realistic testbed for studying overflow in deployed systems. We use test set examples that were \emph{correctly} answered with uncompressed context, filtering out instances where the model fails regardless of compression. This ensures overflow detection focuses on compression-induced failures rather than inherent task difficulty.

\paragraph{Evaluation protocol} Answer correctness is evaluated using GPT-4o-mini for SQuADv2, which assesses semantic equivalence between generated and ground-truth answers. For TriviaQA and HotpotQA, we apply a substring-based exact-match criterion: predictions are marked correct if they contain any reference answer as a substring. All classifiers are evaluated using 5-fold stratified cross-validation; training and hyperparameter details are provided in \S3.5 and Appendix~\ref{app:hyperparameters}.

\subsection{Main Results}

\begin{table*}[t]
  \centering
  \small
  \begin{tabular}{llccc}
  \toprule
  \textbf{Stage} & \textbf{Features} & \textbf{TriviaQA} & \textbf{SQuADv2} & \textbf{HotpotQA} \\
  \midrule
  \multicolumn{1}{l}{\textbf{Pre-compression}} & Context & 0.589 $\pm$ 0.019 & 0.605 $\pm$ 0.025 & 0.541 $\pm$ 0.017 \\
  \midrule
   & Representation & 0.687 $\pm$ 0.015 & 0.662 $\pm$ 0.015 & 0.653 $\pm$ 0.023 \\
   \multicolumn{1}{l}{\textbf{Pre-inference}} & Representation-joint & \textbf{0.725 $\pm$ 0.021} & \underline{0.703 $\pm$ 0.019} & \underline{0.720 $\pm$ 0.007} \\
   & Saturation & 0.568 $\pm$ 0.022 & 0.529 $\pm$ 0.024 & 0.533 $\pm$ 0.010 \\
  \midrule
   & Attention & 0.627 $\pm$ 0.012 & 0.608 $\pm$ 0.020 & 0.623 $\pm$ 0.013 \\
   & Representation & 0.684 $\pm$ 0.016 & 0.665 $\pm$ 0.016 & 0.655 $\pm$ 0.024 \\
   \multicolumn{1}{l}{\textbf{Post-inference}} & Representation-joint & \underline{0.719 $\pm$ 0.015} & \textbf{0.713 $\pm$ 0.016} & \textbf{0.733 $\pm$ 0.011} \\
   & Saturation & 0.583 $\pm$ 0.019 & 0.546 $\pm$ 0.011 & 0.549 $\pm$ 0.008 \\
   & Saturation-joint & 0.600 $\pm$ 0.014 & 0.585 $\pm$ 0.016 & 0.633 $\pm$ 0.015 \\
  \bottomrule
  \end{tabular}
    \caption{Overflow prediction performance (ROC-AUC) across different pipeline stages. \textbf{Pre-compression}: a priori context features. \textbf{Pre-inference}: pre-LLM inference combining preprojection and postprojection features. \textbf{Post-inference}: post-LLM inference combining features from middle and last layers. Representation-joint combines query and context representations. \textbf{Bold}: best performance per dataset; \underline{underlined}: second-best.}
  \label{tab:main_results}  
\end{table*}

% We organize our findings around the three research questions posed in \S1, advancing from characterizing overflow (\textbf{RQ1}) through efficient detection methods (\textbf{RQ2}) to understanding the role of query-context interactions (\textbf{RQ3}). Table~\ref{tab:main_results} presents our main results across three datasets, comparing detection performance at different pipeline stages. Table~\ref{tab:internal_ablation} (Appendix~\ref{app:features_ablation}) provides a detailed ablation study examining feature extraction at different architectural layers.

We organize our findings around the three research questions posed in \S1, advancing from characterizing overflow (\textbf{RQ1}) through efficient detection methods (\textbf{RQ2}) to understanding the role of query-context interactions (\textbf{RQ3}). Table~\ref{tab:main_results} presents our main results across three datasets, comparing detection performance at different pipeline stages. We distinguish between two detection stages: \textbf{pre-inference} (before LLM processing) uses concatenated preprojection and postprojection representations for probing, while \textbf{post-inference} (requiring full LLM forward pass) uses concatenated middle and last layer hidden states. For saturation statistics, the same stage distinction applies, with features extracted from corresponding layer representations. Table~\ref{tab:internal_ablation} (Appendix~\ref{app:features_ablation}) provides a detailed ablation study examining feature extraction at different architectural layers.

\subsubsection{RQ1: Characterizing overflow in compressed representations}

To understand the nature of overflow, we first examined whether compressed tokens exhibit distinctive geometric properties that might correlate with information loss.

\paragraph{Saturation statistics distinguish token types but not overflow} We compared the defined saturation statistics across xRAG and non-xRAG tokens at multiple LLM layers across all three datasets. To avoid positional bias and control for contextual confounds, we compare xRAG token statistics against four baselines: (i) \textit{mean of all non-xRAG tokens} (when compression is applied), capturing the aggregate behavior of standard tokens in compressed sequences; (ii) \textit{mean of original context tokens}, representing uncompressed context behavior; (iii) \textit{first original context token}, isolating position-specific effects; and (iv) \textit{first token in no-context scenarios}, establishing a baseline without any context information.

Tables~\ref{tab:saturation_percentage_squad},~\ref{tab:saturation_percentage_trivia}, and ~\ref{tab:saturation_percentage_hotpot} (Appendix~\ref{app:saturation_stat}) present percentage differences across these baselines. The results reveal consistent patterns across datasets: xRAG tokens show lower sparsity and kurtosis, and dramatically higher spectral entropy across all layers (all $p < 0.001$). Spectral entropy shows the largest differences (87\% across all datasets and baselines), while excess kurtosis shows substantial differences ranging from 29--98\% depending on layer and baseline. Hoyer's sparsity demonstrates more modest but consistent differences of 7--33\%.

Crucially, these patterns remain remarkably stable across datasets and all four baselines, validating that observed properties reflect genuine characteristics of xRAG tokens rather than measurement artifacts or positional biases. The differences persist from middle to final layers, suggesting that compression effects propagate through the network without being normalized away. To verify that these representational differences enable token-type identification, we tested linear separability between xRAG and non-xRAG tokens across all baseline configurations. Linear classifiers achieve near-perfect separation (> 0.95 AUC-ROC for all variants), confirming that saturation statistics reliably distinguish compressed from uncompressed representations in the model's activation space. Notably, more complex classifier architectures (MLP, MLP-SCL) provide no improvement over linear models for this token-type classification task (see Appendix~\ref{app:ablation_study}), further confirming that compressed and uncompressed tokens occupy distinctly separable regions.

However, while these metrics successfully \emph{characterize} compressed tokens, they fail to \emph{predict overflow}. Despite the substantial magnitude of differences and near-perfect linear separability of token types, saturation statistics achieve only near-random predictive performance for overflow detection across datasets (Table~\ref{tab:main_results}). Even when combined with query information (Saturation-joint), performance remains limited (0.55--0.63 AUC-ROC).

\paragraph{Context complexity provides minimal signal} Context-level features (shown in Table~\ref{tab:main_results}) also achieve near-random performance, only marginally exceeding saturation statistics. This indicates that overflow is not strongly predicted by general context properties alone (perplexity, length, statistical compressibility) in our experimental setting. While our current datasets involve relatively short passages compressed into single tokens, we suggest that context complexity features may become more informative in settings with substantially longer contexts or more extreme compression ratios.

\textbf{Summary for RQ1:} Saturation statistics provide a reliable method to separate compressed tokens from uncompressed tokens, achieving near-perfect linear separability and revealing distinct activation-space statistics with 7--87\% relative differences across multiple metrics, layers, and tokens. However, these query-agnostic properties do not predict task-relevant information loss, indicating that while compressed tokens are distinct in representation space, overflow detection requires modeling query-context interactions beyond intrinsic token characteristics.

% \begin{table*}[h!]
%   \centering
%   \small
%   \begin{tabular}{llcccc}
%     \hline
%     Stage & Statistic & Non-xRAG & Context (first) & Context (mean) & No Context \\
%     \hline
%     Middle Layer & Excess Kurtosis & 92.0 & 29.1 & 29.6 & -23.1 \\
%     Middle Layer & Hoyer’s index & 24.4 & 23.0 & 19.4 & 20.1 \\
%     Middle Layer & Spectral Entropy & 0.1 & 87.1 & 87.1 & 87.1 \\
%     Last Layer & Excess Kurtosis & 98.5 & 98.8 & 94.1 & 80.4 \\
%     Last Layer & Hoyer’s index & 31.4 & 32.1 & 19.2 & 7.2 \\
%     Last Layer & Spectral Entropy & 0.1 & 87.1 & 87.1 & 87.0 \\
%     \hline
%   \end{tabular}
%   \caption{Percentage differences in saturation statistics: xRAG vs. non-xRAG tokens. Relative change: $\frac{\text{baseline} - \text{xRAG}}{\text{baseline}} \times 100\%$. Positive values indicate lower saturation (less uniform, more structured). Baselines: mean of all non-xRAG tokens (Non-xRAG), mean of context tokens (Context mean), first context token (Context first), and first no-context token (No Context). Values ($\geq 50\%$) indicate large differences. Consistency across baselines confirms genuine xRAG properties rather than measurement artifacts.}
% \label{tab:saturation_percentage_SQuADv2}
% \end{table*}

\begin{table*}[h!]
  \centering
  \small
  \begin{tabular}{llcccc}
    \hline
    \textbf{Stage} & \textbf{Statistic} & \textbf{Non-xRAG} & \textbf{Context (first)} & \textbf{Context (mean)} & \textbf{No Context} \\
    \hline
    & Excess Kurtosis & 92.0 & 29.1 & 29.6 & -23.1 \\
    \multicolumn{1}{l}{\textbf{Middle Layer}} & Hoyer’s index & 24.4 & 23.0 & 19.4 & 20.1 \\
    & Spectral Entropy & 0.1 & 87.1 & 87.1 & 87.1 \\
    \hline
    & Excess Kurtosis & 98.5 & 98.8 & 94.1 & 80.4 \\
    \multicolumn{1}{l}{\textbf{Last Layer}} & Hoyer’s index & 31.4 & 32.1 & 19.2 & 7.2 \\
    & Spectral Entropy & 0.1 & 87.1 & 87.1 & 87.0 \\
    \hline
  \end{tabular}
  \caption{Relative differences (\%) in saturation statistics between xRAG and baseline tokens on SQuADv2, computed as $\frac{\text{baseline} - \text{xRAG}}{\text{baseline}} \times 100\%$. \textbf{Middle Layer}: LLM intermediate layer features. \textbf{Last Layer}: LLM final layer features. Positive values indicate xRAG tokens have lower saturation (more structured representations). Large differences ($\geq 50\%$) in Excess Kurtosis and Spectral Entropy demonstrate consistent xRAG-specific properties across multiple baselines.}
\label{tab:saturation_percentage_squad}
\end{table*}

\subsubsection{RQ2: Efficient overflow detection without full LLM inference}

While saturation statistics and context complexity features show limited predictive power, we investigated whether learned classifiers can effectively detect overflow, and critically, \emph{at which stage in the compression pipeline} degradation becomes detectable. Tables~\ref{tab:main_results} and~\ref{tab:internal_ablation} compare detection performance at two stages: \textbf{pre-inference} (projection stage, before LLM processing) and \textbf{post-inference} (LLM hidden states, after forward pass).

\paragraph{Overflow is detectable immediately after compression} Learned probing classifiers achieve 0.72 AUC-ROC on average at the post-projection stage (Table~\ref{tab:main_results}), substantially outperforming context-only models and query-agnostic baselines. Crucially, \emph{compression-induced information loss manifests in the representation space immediately after projection}, before any LLM processing. The overflow signal is already present in query-context alignment patterns, revealing that degradation is determined by the compression step itself rather than emerging during generation.

\paragraph{LLM processing provides no additional signal} Post-inference detection using middle-layer hidden states achieves identical performance, confirming that overflow established at compression time merely propagates through the network without amplification or masking (Table~\ref{tab:internal_ablation}). Attention patterns (0.62 AUC-ROC on average) and saturation statistics (even when query-conditioned) provide no meaningful improvement over projection-stage features. Last-layer features show slightly degraded performance, suggesting earlier layers better preserve overflow-relevant signals.

\textbf{Summary for RQ2:} Overflow detection without LLM inference matches post-inference performance. This reveals that compression degradation manifests immediately after projection and is determined during compression rather than during generation, enabling both efficient detection and deeper understanding of compression capacity limits.

\subsubsection{RQ3: The necessity of modeling query-context interactions}

Our final question addresses whether overflow can be detected from compressed tokens alone or whether incorporating query information improves detection performance.

\paragraph{Joint representations substantially outperform single-source models} Tables~\ref{tab:main_results} and~\ref{tab:internal_ablation} compare detection using context-only representations (Representation), joint query-context representations (Representation-joint), and query-agnostic saturation statistics. Given the poor performance of saturation statistics alone, we explored whether incorporating contextual information could improve detection by aggregating statistics (Hoyer's sparsity, spectral entropy, and excess kurtosis) from all non-xRAG tokens in the compressed sequence, computing their mean, maximum, minimum, and standard deviation (Saturation-joint). 

The results reveal a clear hierarchy: representation-joint models achieve 0.70--0.73 AUC-ROC across datasets and stages, substantially outperforming context-only models (0.64--0.69 AUC-ROC). Saturation-joint yields only modest improvements over saturation-only features (0.58--0.63 vs. 0.52--0.58 AUC-ROC), remaining substantially below representation-based methods. This confirms that token-level activation statistics alone are insufficient for overflow detection, regardless of aggregation strategy.

This reveals that overflow is not an intrinsic property of compressed representations but emerges from the \emph{mismatch} between what information the compressed token contains and what the query requires. Joint representation models capture this alignment directly in representation space, enabling accurate overflow prediction. Notably, saturation statistics maintain consistent low performance across all pipeline stages, confirming their utility for \emph{identifying} compressed tokens but not for \emph{predicting} query-specific overflow. Similar to token-type classification (RQ1), linear classifiers prove sufficient for overflow detection, with more complex architectures providing minimal improvement (Appendix~\ref{app:ablation_study}), suggesting that overflow manifests as relatively simple (approximately linearly separable) structure in joint representation space.

\textbf{Summary for RQ3:} Overflow detection fundamentally requires modeling query-context interactions. Joint representation models yield the strongest performance, outperforming context-only models by 5--8 percentage points (Table~\ref{tab:main_results}).

\section{Conclusion}
We investigated token overflow in soft compression architectures and proposed a methodology advancing from query-independent to query-aware detection. Our findings show that saturation statistics reliably separate compressed from uncompressed tokens (7--87\% relative differences), while learned probing on joint query-context representations achieves efficient pre-inference overflow detection (0.72 AUC-ROC on average) without LLM forward passes. Post-inference detection achieves comparable performance, confirming that overflow can be detected efficiently before expensive LLM processing. These results enable safer deployment of compression modules through low-cost pre-LLM gating and adaptive chunking strategies.

\section*{Limitations}
This work focuses on the xRAG architecture as an initial controlled study. Future work should extend the methodology to longer contexts, diverse tasks (summarization, multi-hop reasoning), and other compression architectures to validate generalizability. Exploring richer overflow definitions beyond task performance degradation could capture subtle information loss patterns. Our detection performance establishes a strong baseline, with promising directions including multi-task learning across different compression ratios, incorporating architectural features of the compressor, and developing adaptive systems that dynamically adjust compression based on predicted overflow risk. The methodology's architecture-agnostic design facilitates such extensions to emerging compression techniques.

\section*{Ethical Considerations} Generation of text with LLMs using compressed and overflown tokens can lead to hallucinations and  untrustworthy output. This way, the created technology may be considered helpful for minimizing such effects. At the same time, as the absolute accuracy numbers of the developed classifier are relatively low, and eventual false positive predictions could lead to overconfidence in trustworthiness of generated texts. Therefore, we suggest that more research is needed to raise the absolute values of the developed classifiers to ensure their safe use in various text generation workflows and applications.

% \section*{Acknowledgements}

% The work of Alexander Panchenko was supported by the RSF project № 25-71-30008 ``Laboratory for reliable, adaptive, and trustworthy Artificial Intelligence''.

\bibliography{custom}

\onecolumn
\appendix
\section{Classifiers Ablation Study}\label{app:ablation_study}
\begin{figure*}[h!]
\centering
\includegraphics[width=\textwidth]{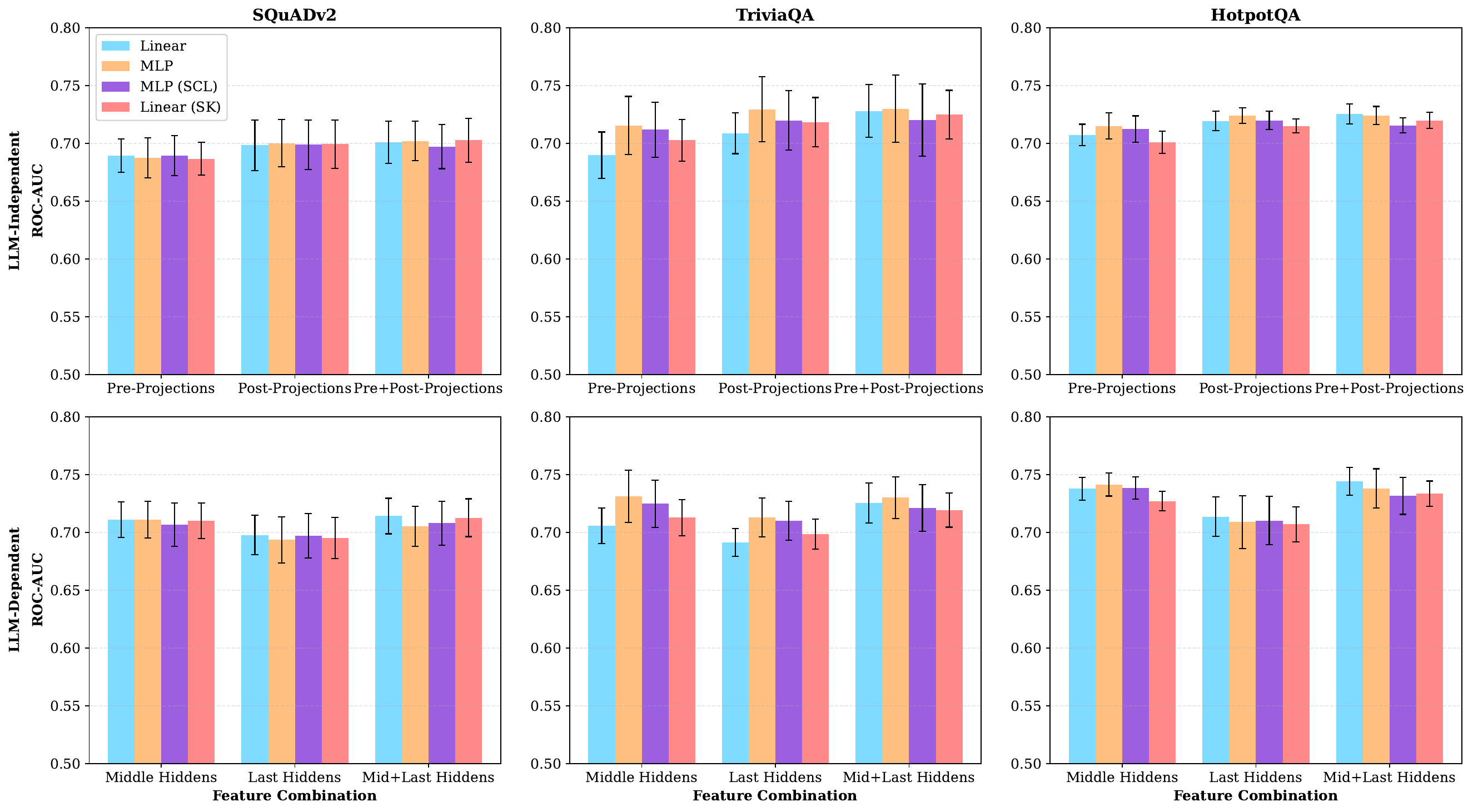}
\caption{Comparison of classifier architectures (Linear scikit-learn, Linear PyTorch, MLP, MLP with SCL) across datasets and feature combinations. All architectures achieve comparable performance, with differences typically $<$1 percentage point, demonstrating that overflow is largely linearly separable in joint representation space.}
\label{fig:classifier_comparison}
\end{figure*}

\newpage
\section{Features Ablation Study}\label{app:features_ablation}
\begin{table*}[h!]
\centering
\small
\begin{tabular}{llccc}
\toprule
\textbf{Stage} & \textbf{Features} & \textbf{TriviaQA} & \textbf{SQuADv2} & \textbf{HotpotQA} \\
\midrule
\multicolumn{1}{l}{\textbf{Pre-projection}} & Representation & 0.679 $\pm$ 0.019 & 0.641 $\pm$ 0.013 & 0.635 $\pm$ 0.017 \\
 & Representation-joint & 0.703 $\pm$ 0.018 & 0.687 $\pm$ 0.014 & 0.701 $\pm$ 0.010 \\
 & Saturation & 0.534 $\pm$ 0.016 & 0.526 $\pm$ 0.024 & 0.522 $\pm$ 0.009 \\
\midrule
\multicolumn{1}{l}{\textbf{Post-projection}} & Representation & 0.675 $\pm$ 0.016 & 0.660 $\pm$ 0.016 & 0.652 $\pm$ 0.025 \\
 & Representation-joint & \textbf{0.718 $\pm$ 0.021} & \underline{0.699 $\pm$ 0.021} & \underline{0.715 $\pm$ 0.006} \\
 & Saturation & 0.553 $\pm$ 0.017 & 0.530 $\pm$ 0.011 & 0.518 $\pm$ 0.011 \\
\midrule
\multicolumn{1}{l}{\textbf{Middle layer}} & Attention & 0.620 $\pm$ 0.012 & 0.581 $\pm$ 0.008 & 0.603 $\pm$ 0.014 \\
 & Representation & 0.675 $\pm$ 0.016 & 0.660 $\pm$ 0.016 & 0.652 $\pm$ 0.025 \\
 & Representation-joint & \underline{0.713 $\pm$ 0.015} & \textbf{0.710 $\pm$ 0.015} & \textbf{0.727 $\pm$ 0.009} \\
 & Saturation & 0.553 $\pm$ 0.017 & 0.530 $\pm$ 0.011 & 0.518 $\pm$ 0.011 \\
 & Saturation-joint & 0.557 $\pm$ 0.021 & 0.560 $\pm$ 0.026 & 0.597 $\pm$ 0.017 \\
\midrule
\multicolumn{1}{l}{\textbf{Last layer}} & Attention & 0.584 $\pm$ 0.011 & 0.584 $\pm$ 0.027 & 0.615 $\pm$ 0.014 \\
 & Representation & 0.675 $\pm$ 0.016 & 0.661 $\pm$ 0.016 & 0.653 $\pm$ 0.024 \\
 & Representation-joint & 0.698 $\pm$ 0.013 & 0.695 $\pm$ 0.018 & 0.707 $\pm$ 0.015 \\
 & Saturation & 0.554 $\pm$ 0.020 & 0.523 $\pm$ 0.015 & 0.530 $\pm$ 0.010 \\
 & Saturation-joint & 0.582 $\pm$ 0.015 & 0.541 $\pm$ 0.011 & 0.618 $\pm$ 0.011 \\
\bottomrule
\end{tabular}
\caption{Ablation study examining feature extraction at different architectural stages (ROC-AUC). \textbf{Pre-projection}: retriever embeddings before projection. \textbf{Post-projection}: representations after projection. \textbf{Middle layer}: intermediate LLM hidden states. \textbf{Last layer}: final LLM hidden states. Representation-joint combines query and context representations. \textbf{Bold}: best performance per dataset; \underline{underlined}: second-best.}

\label{tab:internal_ablation}
\end{table*}

\newpage
\section{Saturation Statistics}\label{app:saturation_stat}
% \begin{table*}[h!]
%   \centering
%   \small
%   \begin{tabular}{llcccc}
%     \hline
%     Stage & Statistic & Non-xRAG & Context (first) & Context (mean) & No Context \\
%     \hline
%     Middle Layer & Excess Kurtosis & 92.4 & 41.3 & 40.5 & -4.4 \\
%     Middle Layer & Hoyer’s index & 24.4 & 26.1 & 20.6 & 21.1 \\
%     Middle Layer & Spectral Entropy & 0.1 & 87.1 & 87.1 & 87.1 \\
%     Last Layer & Excess Kurtosis & 98.3 & 98.4 & 93.1 & 81.4 \\
%     Last Layer & Hoyer’s index & 28.4 & 28.1 & 16.0 & 7.4 \\
%     Last Layer & Spectral Entropy & 0.0 & 87.1 & 87.1 & 87.0 \\
%     \hline
%   \end{tabular}
%   \caption{Percentage differences in saturation statistics: xRAG vs. non-xRAG tokens. Relative change: $\frac{\text{baseline} - \text{xRAG}}{\text{baseline}} \times 100\%$. Positive values indicate lower saturation (less uniform, more structured). Baselines: mean of all non-xRAG tokens (Non-xRAG), mean of context tokens (Context mean), first context token (Context first), and first no-context token (No Context). Values ($\geq 50\%$) indicate large differences. Consistency across baselines confirms genuine xRAG properties rather than measurement artifacts.}
% \label{tab:saturation_percentage_trivia}
% \end{table*}

\begin{table*}[h!]
  \centering
  \small
  \begin{tabular}{llcccc}
    \hline
    \textbf{Stage} & \textbf{Statistic} & \textbf{Non-xRAG} & \textbf{Context (first)} & \textbf{Context (mean)} & \textbf{No Context} \\
    \hline
    & Excess Kurtosis & 92.4 & 41.3 & 40.5 & -4.4 \\
    \multicolumn{1}{l}{\textbf{Middle Layer}} & Hoyer’s index & 24.4 & 26.1 & 20.6 & 21.1 \\
    & Spectral Entropy & 0.1 & 87.1 & 87.1 & 87.1 \\
    \hline
    & Excess Kurtosis & 98.3 & 98.4 & 93.1 & 81.4 \\
    \multicolumn{1}{l}{\textbf{Last Layer}} & Hoyer’s index & 28.4 & 28.1 & 16.0 & 7.4 \\
    & Spectral Entropy & 0.0 & 87.1 & 87.1 & 87.0 \\
    \hline
  \end{tabular}
  \caption{Relative differences (\%) in saturation statistics between xRAG and baseline tokens on TriviaQA, computed as $\frac{\text{baseline} - \text{xRAG}}{\text{baseline}} \times 100\%$. \textbf{Middle Layer}: LLM intermediate layer features. \textbf{Last Layer}: LLM final layer features. Positive values indicate xRAG tokens have lower saturation (more structured representations). Large differences ($\geq 50\%$) in Excess Kurtosis and Spectral Entropy demonstrate consistent xRAG-specific properties across multiple baselines.}
\label{tab:saturation_percentage_trivia}
\end{table*}
% \begin{table*}[h!]
%   \centering
%   \small
%   \begin{tabular}{llcccc}
%     \hline
%     Stage & Statistic & Non-xRAG & Context (first) & Context (mean) & No Context \\
%     \hline
%     Middle Layer & Excess Kurtosis & 91.6 & 39.7 & 46.0 & -10.6 \\
%     Middle Layer & Hoyer’s index & 24.4 & 25.1 & 22.0 & 20.8 \\
%     Middle Layer & Spec Entropy & 0.1 & 87.1 & 87.1 & 87.1 \\
%     Last Layer & Excess Kurtosis & 98.1 & 98.9 & 92.3 & 80.8 \\
%     Last Layer & Hoyer’s index & 28.3 & 33.5 & 14.7 & 7.0 \\
%     Last Layer & Spec Entropy & 0.0 & 87.1 & 87.1 & 87.0 \\
%     \hline
%   \end{tabular}
%   \caption{Percentage differences in saturation statistics: xRAG vs. non-xRAG tokens. Relative change: $\frac{\text{baseline} - \text{xRAG}}{\text{baseline}} \times 100\%$. Positive values indicate lower saturation (less uniform, more structured). Baselines: mean of all non-xRAG tokens (Non-xRAG), mean of context tokens (Context mean), first context token (Context first), and first no-context token (No Context). Values ($\geq 50\%$) indicate large differences. Consistency across baselines confirms genuine xRAG properties rather than measurement artifacts.}
% \label{tab:saturation_percentage_hotpot}
% \end{table*}

\begin{table*}[h!]
  \centering
  \small
  \begin{tabular}{llcccc}
    \hline
    \textbf{Stage} & \textbf{Statistic} & \textbf{Non-xRAG} & \textbf{Context (first)} & \textbf{Context (mean)} & \textbf{No Context} \\
    \hline
    & Excess Kurtosis & 91.6 & 39.7 & 46.0 & -10.6 \\
    \multicolumn{1}{l}{\textbf{Middle Layer}} & Hoyer’s index & 24.4 & 25.1 & 22.0 & 20.8 \\
    & Spectral Entropy & 0.1 & 87.1 & 87.1 & 87.1 \\
    \hline
    & Excess Kurtosis & 98.1 & 98.9 & 92.3 & 80.8 \\
    \multicolumn{1}{l}{\textbf{Last Layer}} & Hoyer’s index & 28.3 & 33.5 & 14.7 & 7.0 \\
    & Spectral Entropy & 0.0 & 87.1 & 87.1 & 87.0 \\
    \hline
  \end{tabular}
  \caption{Relative differences (\%) in saturation statistics between xRAG and baseline tokens on HotpotQA, computed as $\frac{\text{baseline} - \text{xRAG}}{\text{baseline}} \times 100\%$. \textbf{Middle Layer}: LLM intermediate layer features. \textbf{Last Layer}: LLM final layer features. Positive values indicate xRAG tokens have lower saturation (more structured representations). Large differences ($\geq 50\%$) in Excess Kurtosis and Spectral Entropy demonstrate consistent xRAG-specific properties across multiple baselines.}
\label{tab:saturation_percentage_hotpot}
\end{table*}

\newpage
\section{Hyperparameters}\label{app:hyperparameters}

Table~\ref{tab:hyperparameters} summarizes all hyperparameters used in our experiments. All hyperparameters were tuned on the SQuADv2 validation set and then fixed across TriviaQA and HotpotQA datasets.

\begin{table}[h!]
\centering
\small
\begin{tabular}{lll}
\toprule
\textbf{Method} & \textbf{Parameter} & \textbf{Value} \\
\midrule
\multicolumn{3}{@{}l}{\textit{Feature-based Classification (Logistic Regression)}} \\
& Solver & lbfgs \\
& Regularization & L2, $C = 10^{-5}$ \\
& Max iterations & 1000 \\
& Preprocessing & StandardScaler \\
\midrule
\multicolumn{3}{@{}l}{\textit{Common Settings (All Neural Probes)}} \\
& Cross-validation & 5-fold stratified, 80\%/20\% train/val \\
& Batch size & 256 \\
& Optimizer & Adam, learning rate $= 10^{-4}$ \\
& Preprocessing & StandardScaler \\
& Regularization & $\mathcal{L}_{\text{reg}} = \frac{\lambda_2}{2N}\|\theta\|_2^2 + \frac{\lambda_1}{N}\|\theta\|_1$ \\
& & with $(\lambda_2, \lambda_1) = (500, 100)$ \\
& Early stopping & Patience = 20 epochs \\
\midrule
\multicolumn{3}{@{}l}{\textit{Linear Probe}} \\
& Architecture & Single linear layer \\
& Max epochs & 150 \\
\midrule
\multicolumn{3}{@{}l}{\textit{MLP Probe}} \\
& Architecture & Two-layer feedforward \\
& Hidden dimension & 1024 \\
& Activation & ReLU \\
& Max epochs & 50 \\
\midrule
\multicolumn{3}{@{}l}{\textit{MLP-SCL Probe}} \\
& Architecture & Two-layer feedforward \\
& Hidden dimension & 1024 \\
& Activation & SiLU \\
& Dropout & 0.1 (before and after hidden layer) \\
& Normalization & BatchNorm1d after hidden layer \\
& Contrastive weight & $\lambda = 0.3$ \\
& Temperature & $\tau = 0.07$ \\
& Max epochs & 50 \\
\bottomrule
\end{tabular}
\caption{Hyperparameters for all overflow detection methods. The regularization term for neural probes combines L2 and L1 penalties scaled by the number of model parameters $N$ (excluding biases).}
\label{tab:hyperparameters}
\end{table}

\end{document}